\documentclass[runningheads]{llncs}
\usepackage{amsmath}
\usepackage{amsfonts}
\usepackage{graphicx}
\usepackage{multirow}
\usepackage{pifont}
\usepackage{marvosym}

\begin{document}
\title{DecoratingFusion: A LiDAR-Camera Fusion Network with the Combination of Point-level and Feature-level Fusion}
%\author{First Author\inst{1}\orcidID{0000-1111-2222-3333} \and
%Second Author\inst{2,3}\orcidID{1111-2222-3333-4444} \and
%Third Author\inst{3}\orcidID{2222--3333-4444-5555}}
\author{Zixuan Yin\inst{1} \and Han Sun\inst{1}\textsuperscript{\Letter} \and Ningzhong Liu\inst{1} \and Huiyu Zhou\inst{2} \and Jiaquan Shen\inst{3}}
\authorrunning{Z. Yin et al.}
%\institute{Princeton University, Princeton NJ 08544, USA \and
%Springer Heidelberg, Tiergartenstr. 17, 69121 Heidelberg, Germany
\institute{Nanjing University of Aeronautics and
Astronautics, Nanjing, China 
\email{sunhan@nuaa.edu.cn}
\and University of Leicester, UK \and Luoyang Normal University, Luoyang, China}

\maketitle              % typeset the header of the contribution
\begin{abstract}
Lidars and cameras play essential roles in autonomous driving, offering complementary information for 3D detection. The state-of-the-art fusion methods integrate them at the feature level, but they mostly rely on the learned soft association between point clouds and images, which lacks interpretability and neglects the hard association between them. In this paper, we combine feature-level fusion with point-level fusion, using hard association established by the calibration matrices to guide the generation of object queries. Specifically, in the early fusion stage, we use the 2D CNN features of images to decorate the point cloud data, and employ two independent sparse convolutions to extract the decorated point cloud features. In the mid-level fusion stage, we initialize the queries with a center heatmap and embed the predicted class labels as auxiliary information into the queries, making the initial positions closer to the actual centers of the targets. Extensive experiments conducted on two popular datasets, i.e. KITTI, Waymo, demonstrate the superiority of DecoratingFusion.

\end{abstract}

\section{Introduction}
In recent years, lidar-camera fusion methods have been increasingly applied in 3D object detection for autonomous driving scenarios\cite{zhang2022cat}. Point cloud data provides spatial geometric information, describing object shape, position, and size, while image data provides color, texture, and visual features. However, due to significant domain gaps between the two modalities, feature alignment has become a key challenge.

After undergoing the development of late-fusion at the result level and early-fusion at the point level, the current state-of-the-art fusion method is mid-level feature fusion\cite{zhou2022centerformer}. This kind of method maximizes the complementarity between the two types of data but is also the most challenging to implement. Due to the high abstraction level of the features, the key challenge lies in how to match the data from the two modalities in the feature space effectively.

Existing feature fusion methods mostly rely on learning-based approaches to obtain soft correlations between point clouds and images. However, these soft correlations lack interpretability, making it difficult to ensure their reliability. In contrast, the hard correlations commonly used in point-level fusion methods (such as calibration matrices between LiDAR and cameras, which are usually provided by data collectors) are intuitive, ensuring the alignment between point clouds and images. However, hard correlations are rarely applied in feature fusion methods because they become feature-to-feature mappings in the feature space, rendering the use of hard correlations impractical.

To introduce hard correlations into feature fusion methods, we combine early fusion and mid-level fusion and propose an efficient multi-modal fusion strategy called DecoratingFusion. Specifically, we utilize a 2D CNN to extract image features and employ them to decorate the original point cloud data. These convolutional layers, along with other network components, are trained in an end-to-end manner. Considering the domain gaps between the two modalities, two independent sparse convolutions are used to extract the decorated point cloud features, which are then concatenated. These concatenated features are utilized to generate object queries and fuse point cloud and image features through cross-attention mechanisms. Additionally, to optimize the initialization process of object queries, we first predict a center heatmap using the decorated point cloud features and select the initialization position of the query from the heatmap. Finally, the fused features are fed into the prediction head to obtain the final results.

In brief, our contributions can be summarized as follows:
\begin{itemize}
    \item We combine the early fusion and mid-level fusion approaches, utilizing the hard correlations established through calibration matrices to guide the generation of object queries, and fuse the point cloud and image features using cross-attention mechanisms.
    \item We use two independent sparse convolutions to extract the decorated point cloud features. Additionally, we initialize the object queries using a center heatmap and embed the predicted class from the center heatmap as auxiliary information into the object queries.
    \item We validate our method on two prominent autonomous driving datasets, namely KITTI and Waymo. The experimental results demonstrate the effectiveness of our proposed approach.
\end{itemize}

\section{Related Work}
According to the fusion stage, existing fusion methods can be divided into three categories: early fusion, mid-level fusion, and late fusion. Below, we will introduce each of them in chronological order of their development.

\subsection{Late Fusion}
Late fusion methods refer to fusion performed at the later stages of the network, also known as result-level fusion. Early multi-modal fusion methods are mostly late fusion methods, where each modality's data is independently predicted, and the detection results are subsequently fused. This category of methods has the simplest fusion approach and high reliability because they can still operate normally under single-modality conditions. However, their fusion granularity is the lowest among the three categories, often resulting in poor performance. Classic late fusion methods include F-PointNets\cite{qi2018frustum}, CLOCs\cite{pang2020clocs}. In recent years, early fusion methods and mid-level fusion methods have gained more attention, while research on late fusion methods has been relatively limited.

\subsection{Early Fusion}
Early fusion methods refer to fusion performed at the early stages of the network, where different modalities of data are fused to create a new modality, which serves as the input to the feature extraction network. For example, PointPainting\cite{vora2020pointpainting} utilizes pixel-level semantic segmentation scores to decorate point cloud data as camera features, while PointAugmenting\cite{wang2021pointaugmenting} decorates point cloud data with image features. Although PointPainting and PointAugmenting provide novel approaches for multi-modal fusion, researchers quickly discover the bottleneck in them. While these fusion approach allow for the alignment of 3D and 2D coordinates using calibration matrices, existing feature extractors struggle to directly process the fused data. Typical 3D backbone networks are designed specifically to extract sparse point cloud features, and as point cloud and image data have significant differences in characteristics, using existing 3D backbone networks to extract features from the fused data is not suitable.

\subsection{Mid-level Fusion}
Mid-level fusion methods refer to fusion performed during the feature extraction stage of the network. Compared to early fusion, which can only fuse data at the raw data level, mid-level fusion methods can operate at the feature level. Therefore, in theory, mid-level fusion methods can maximize the advantages of fusion. However, most existing mid-level fusion methods rely on learned soft correlations between point clouds and images. We consider this approach to be unreliable because it lacks interpretability, and existing feature-level fusion methods often overlook the presence of hard correlations. For example, TransFusion\cite{bai2022transfusion} introduces the SMCA (Spatially Modulated Cross Attention) module, which allows the network to adaptively determine which parts of the image features are more important and suitable for fusion. DeepFusion\cite{li2022deepfusion} enables a voxel in the point cloud to match multiple pixels in the image and assigns weights to the pixels using cross-attention mechanisms.

\section{DecoratingFusion}
\subsection{Motivations and Pipeline}
Before our work, early fusion methods and mid-level fusion methods are completely different approaches. We aimed to leverage the advantages of both by incorporating the hard correlations established through calibration matrices to achieve better detection results. To utilize the hard correlations between point clouds and images to guide the fusion of the two modalities at the feature level, we employ 2D CNN features of correspondingly matched pixels to decorate the original point cloud data. We use the decorated point cloud features to initialize the object queries, aiding the cross-attention module in capturing the correlations between the two types of features more effectively.

\begin{figure}
\includegraphics[width=\textwidth]{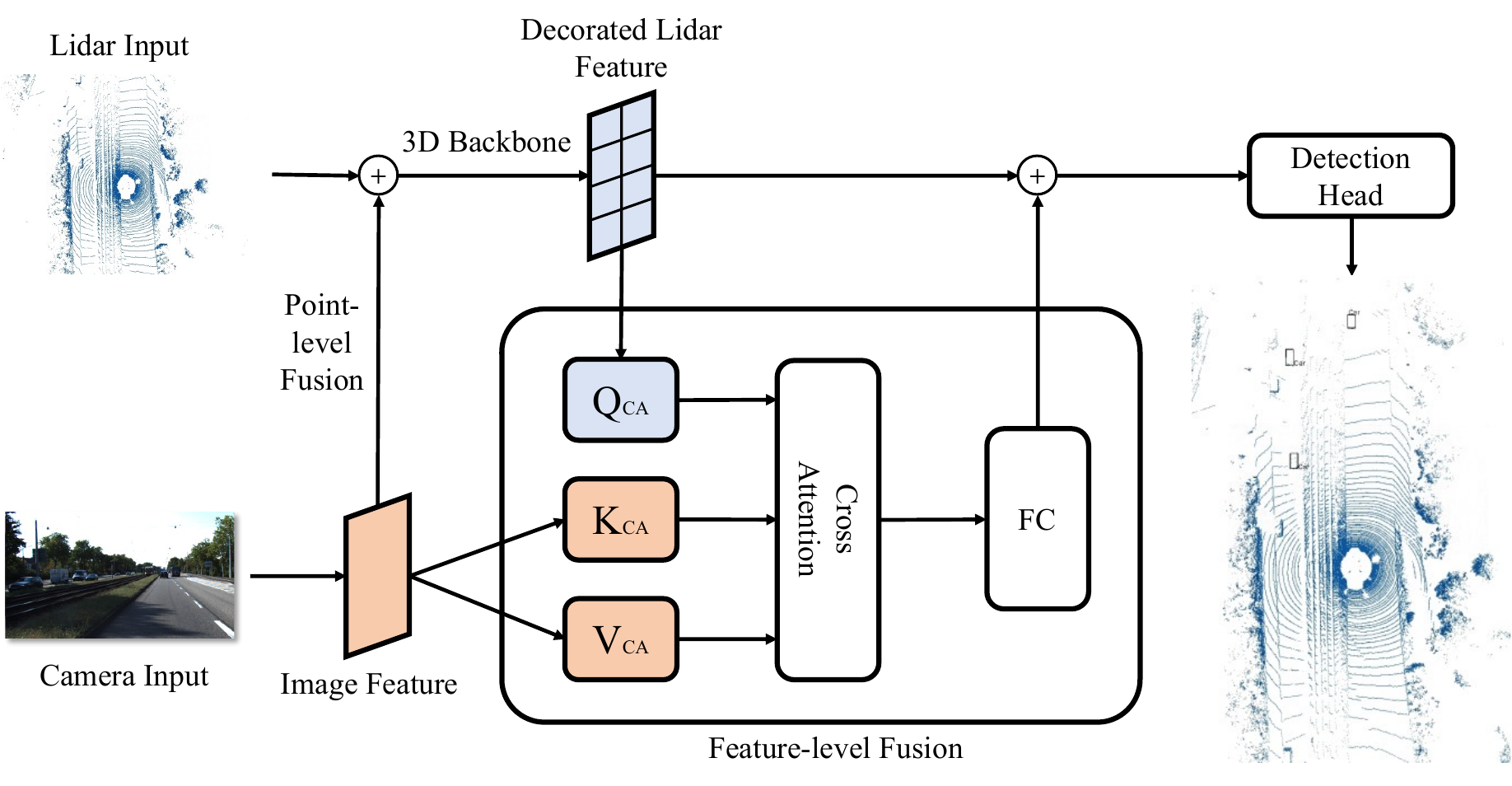}
\caption{An overview of DecoratingFusion framework. } \label{fig_backbone}
\end{figure}

DecoratingFusion consists of two parts: the point-level fusion stage and the feature-level fusion stage, as illustrated in Fig.~\ref{fig_backbone}. First, in the point-level fusion stage, we use a 2D backbone network to extract features from the input image. Then, using calibration matrices, we project the points of the point cloud onto the image plane, obtaining the corresponding pixel points. We attach the image features of those pixel points to the original point cloud. Next, in the feature fusion stage, the decorated point cloud is passed through a 3D backbone network to obtain the decorated lidar features, which are used to generate object queries. Simultaneously, the image features are used to derive keys and values. Then, through cross-attention, we learn the soft correlations between the two types of features. Finally, the fused features are obtained by connecting the point cloud features with the learned soft correlations. Lastly, the fused features are fed into the existing prediction heads to obtain the final detection results.

\subsection{Point-level Fusion}
DecoratingFusion utilizes DLA34 from CenterNet\cite{zhou2019objects} as the 2D backbone network, which produces feature maps with a channel size of 64 and a scale factor of 4. We represent the original point cloud points as $(x,y,z,r)$, where $x$, $y$, $z$ represent the coordinates of the point in 3D space, and $r$ represents the reflectance. Each point can be projected onto the image using the calibration matrix $T_{\text{camera}\leftarrow \text{lidar}}$. In this stage, the decorated point cloud, denoted as $(x,y,z,r,f)$, is obtained, where $f$ represents the image features attached to each point.

In addition, the 2D backbone network of DecoratingFusion is trained end-to-end with other network components, unlike PointPainting\cite{vora2020pointpainting} or PointAugmenting\cite{wang2021pointaugmenting}, which are independently learned in other tasks such as 2D semantic segmentation or object detection. This approach reduces computational costs, mitigates cross-domain differences, decreases the amount of required data annotation, and avoids suboptimal feature extraction due to heuristic feature selection.

\subsection{Feature-level Fusion}
\begin{figure}
\includegraphics[width=\textwidth]{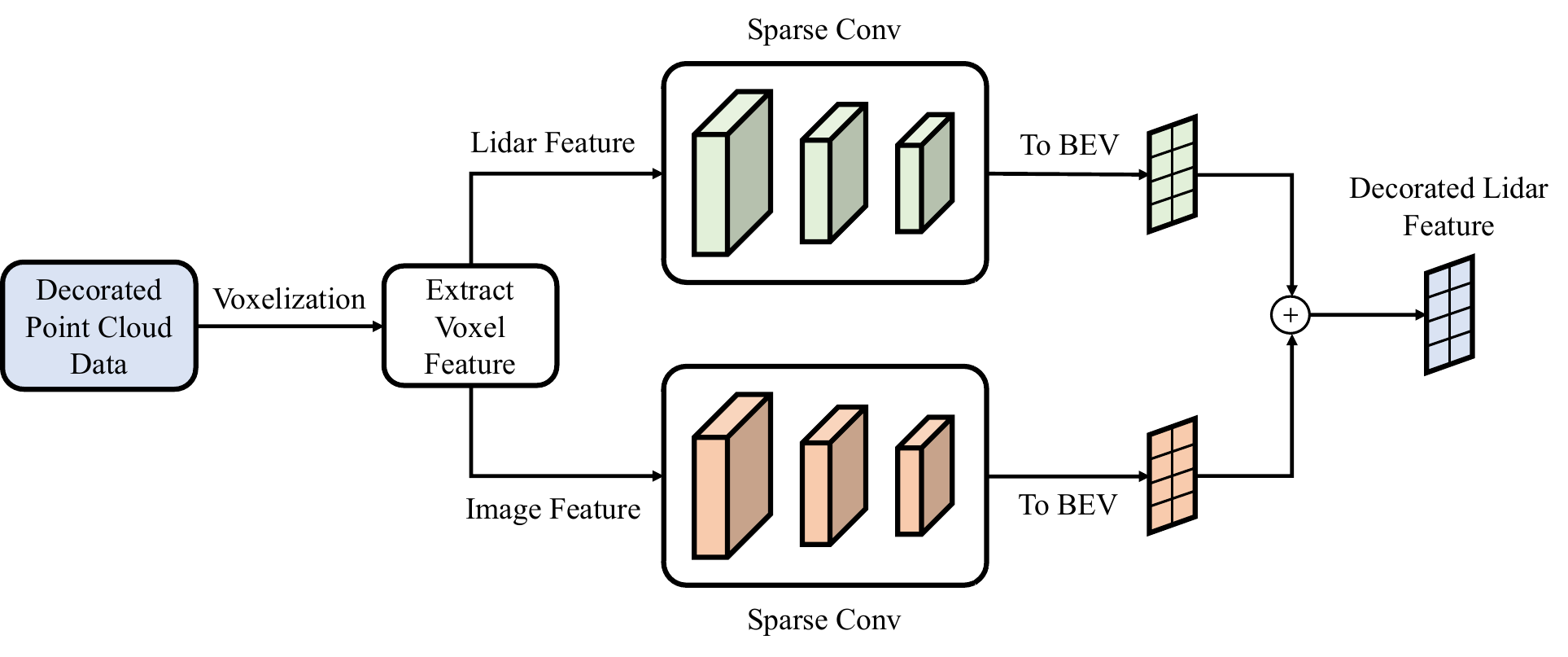}
\caption{The two independent sparse convolutions used to extract lidar and image feature. } \label{fig_2sparseconv}
\end{figure}

After obtaining the decorated point cloud data, considering the domain gap between the two modalities, we do not directly feed them into the existing 3D backbone network. Instead, we process the two types of features separately. Specifically, as shown in Fig.~\ref{fig_2sparseconv}, we first voxelate the decorated point cloud data and extract local features for each voxel. Then, we split the point cloud features and image features, and perform further feature extraction using two independent 3D sparse convolutions. Finally, we compress both modalities' features into a BEV representation and concatenate them along the channel dimension, resulting in the final augmented point cloud features.

With the point cloud features $F^\text{Lidar}$ and image features $F^\text{Camera}$, we can transform $F^\text{Lidar}$ into queries $Q_{\text{CA}}$ and $F^\text{Camera}$ into keys $K_{\text{CA}}$ and values $V_{\text{CA}}$. Then, we use cross-attention to learn the correlations between the two types of features. However, unlike DeepFusion\cite{li2022deepfusion}, which directly uses fully connected layers to transform $F^\text{Lidar}$ into queries $Q_{\text{CA}}$, we are inspired by CenterPoint\cite{yin2021center} and use a center point heatmap for better initialization of object queries. Specifically, we first predict a center point heatmap $\hat{Y}\in \mathbb{R}^{X\times Y\times K}$, where $X\times Y$ represents the size of $F^\text{\text{Lidar}}$ and $K$ represents the number of classes. We treat this heatmap as $X\times Y\times K$ candidate objects and select the top $n$ per class as initial object queries. To prevent queries from being too densely concentrated in a local region, we choose the local maxima as queries, which means their values must be greater than or equal to their eight neighboring points. The query initialization method of DecoratingFusion has the following advantages compared to DeepFusion\cite{li2022deepfusion}: (1) The initial positions of the queries are closer to the actual center positions of the objects. (2) The initial positions of the queries are no longer randomly generated, but related to the input data, which can accelerate the convergence speed of the model.

Additionally, inspired by TransFusion\cite{bai2022transfusion}, we also incorporate class information into each object query. Since the decorated point cloud features are in the BEV space, where object scales are absolute, the scale differences between objects of the same class are minimal. Taking advantage of this characteristic in the BEV space, we encode the class of each query obtained from the center heatmap as a one-hot encoding and concatenate it with the query $Q_{\text{CA}}$. This provides additional class information to assist the cross-attention module, allowing it to focus more on intra-class variations.

Finally, by performing an inner product operation between the query $Q_{\text{CA}}$ and the key $K_{\text{CA}}$, we obtain a correlation matrix that captures the relationship between the point cloud features and image features. After applying softmax normalization, this correlation matrix is used to weight and sum the values $V_{\text{CA}}$, resulting in image features that are relevant to the queries. These image features are then processed through a fully connected layer and concatenated with the decorated point cloud features. The concatenated features are fed into the existing 3D detection head to obtain the final detection results.

\subsection{Loss Function}
The loss function of DecoratingFusion consists of a classification loss $\mathcal{L}_{\text{cls}}$, and a regression loss $\mathcal{L}_{\text{reg}}$. The classification loss is calculated using focal loss, while the regression loss is calculated using Smooth-L1 loss:

\begin{equation}
\label{equ:loss_function}
\mathcal{L}=\mathcal{L}_{\text{cls}}+w\mathcal{L}_{\text{reg}}
\end{equation}
where $w$ is set to 2, following the empirical settings of SECOND\cite{yan2018second}.

\section{Experiments}
To evaluate the effectiveness of DecoratingFusion, we conduct experiments on two commonly used outdoor autonomous driving datasets: the KITTI dataset and the Waymo dataset.

\subsection{Datasets}
The KITTI dataset is one of the most commonly used datasets in the field of autonomous driving before 2020. It consists of 7481 training samples and 7518 testing samples from 3D scenes in autonomous driving. Following the convention, we divide the training data into a training set with 3712 samples and a validation set with 3769 samples. In accordance with the requirements of the KITTI object detection benchmark, we conduct experiments on three categories: cars, pedestrians, and cyclists, and evaluate the results using the average precision with an IoU threshold of 0.7.

In 2020, Waymo released a training dataset called WOD (Waymo OpenDataset) for autonomous driving. It consists of 798 training sequences, 202 validation sequences, and 150 testing sequences. Each sequence contains approximately 200 frames, which include lidar points, camera images, and labeled 3D bounding boxes. We evaluate the performance of different models using the official metrics, AP and APH. We report the results for the LEVEL1 (L1) and LEVEL2 (L2) difficulty levels. LEVEL1 is used for anchor boxes with more than 5 lidar points, while LEVEL2 is used for anchor boxes with at least one lidar point.

\subsection{Implementation Details}
We implement DecoratingFusion using the open-source MMDetection3D framework in PyTorch. For the KITTI dataset, we set the voxel sizes to (0.05m, 0.05m, 0.1m). Since the KITTI dataset only provides annotations from the front camera's perspective, the detection ranges on the X, Y, and Z axes are respectively set as [0, 70.4m], [-40m, 40m], and [-3m, 1m]. For the Waymo dataset, we set the voxel sizes to (0.1m, 0.1m, 0.15m). The detection range on the X and Y axes is set as [-75.2m, 75.2m], and the detection range on the Z axis is set as [-2m, 4m]. During the training process, we utilize the AdamW optimizer with a momentum range of 0.85 to 0.95. The learning rate is adjusted using the one-cycle policy. For both the KITTI and Waymo datasets, the initial learning rates are set to $2e-3$ and $3e-3$, respectively, with a weight decay coefficient of 0.01.

We use a pre-trained CenterNet\cite{zhou2019objects} with DLA34 as the 2D backbone network, with the image size set to $448\times 800$. For the 3D backbone network, we utilize SECOND\cite{yan2018second}. In the Cross Attention Module, we apply a dropout strategy to the correlation matrix to prevent overfitting, with a dropout rate of 0.3. The subsequent fully connected layer has 192 filters. Additionally, we employ the GT-Paste strategy for data augmentation during training. This strategy aids in the convergence of the network but may disrupt the true data distribution. Therefore, following PointAugmenting\cite{wang2021pointaugmenting}, we use a fading strategy during training. Specifically, we apply the GT-Paste data augmentation strategy throughout the early stages of training but disable it in the final 5 epochs, allowing the network to better adapt to the real data distribution.

\subsection{Experimental Results and Analysis}
\begin{table}
\caption{Performance comparison on the KITTI \emph{val} set with AP calculated by 40 recall positions.}
\begin{tabular}{c|c|c|c|c|c|c|c|c|c|c}
\hline
\multirow{2}*{Method} & \multirow{2}*{mAP} & \multicolumn{3}{c|}{Car} & \multicolumn{3}{c|}{Cyclist} & \multicolumn{3}{c}{Pedestrian}\\
\cline{3-11}
& & Easy & Mod. & Hard & Easy & Mod. & Hard & Easy & Mod. & Hard\\
\hline %[2pt]
SECOND\cite{yan2018second} & 68.06 & 88.61 & 78.62 & 77.22 & 80.58 & 67.15 & 63.10 & 56.55 & 52.98 & 47.73 \\
PointRCNN\cite{shi2019pointrcnn} & 70.67 & 88.72 & 78.61 & 77.82 & 86.84 & 71.62 & 65.59 & 62.72 & 53.85 & 50.25 \\
PV-RCNN\cite{shi2020pv} & 73.27 & 92.10 & 84.36 & 82.48 & \underline{88.88} & 71.95 & 66.78 & 64.26 & 56.67 & 51.91 \\
SE-SSD\cite{zheng2021se} & - & 90.21 & \textbf{86.25} & 79.22 & - & - & - & - & - & -\\
\hline
F-PointNet\cite{qi2018frustum} & 65.58 & 83.76 & 70.92 & 63.65 & 77.15 & 56.49 & 53.37 & 70.00 & 61.32 & 53.59 \\
CLOCs\cite{pang2020clocs} & 70.5 & 89.49 & 79.31 & 77.36 & 87.57 & 67.92 & 63.67 & 62.88 & 56.2 & 50.1\\
EPNet\cite{huang2020epnet} & 70.97 & 88.76 & 78.65 & 78.32 & 83.88 & 65.60 & 62.70 & 66.74 & 59.29 & 54.82 \\
FocalsConv\cite{chen2022focal} & - & \textbf{92.26} & \underline{85.32} & \underline{82.95} & - & - & - & - & - & -\\
CAT-Det\cite{zhang2022cat} & \underline{75.42} & 90.12 & 81.46 & 79.15 & 87.64 & \underline{72.82} & \underline{68.20} & \textbf{74.08} & \underline{66.35} & \underline{58.92}\\
\hline
DecoratingFusion & \textbf{77.30} & \underline{92.25} & 85.04 & \textbf{83.82} & \textbf{90.41} & \textbf{74.24} & \textbf{70.51} & \underline{73.22} &\textbf{66.41} & \textbf{59.57}\\
\hline
\end{tabular}
\end{table}

\subsubsection{KITTI.}
To demonstrate the effectiveness of DecoratingFusion, we compare it with nine representative 3D object detection methods on the KITTI dataset. The selected methods include SECOND\cite{yan2018second}, PointRCNN\cite{shi2019pointrcnn}, PVRCNN\cite{shi2020pv}, SE-SSD\cite{zheng2021se} (four lidar-only methods), as well as F-PointNet\cite{qi2018frustum}, CLOCs\cite{pang2020clocs}, EPNet\cite{huang2020epnet}, FocalsConv\cite{chen2022focal}, and CAT-Det\cite{zhang2022cat} (five multi-modal fusion methods). The experimental results are shown in Table 1, where the top-ranking score is displayed in bold, and the second-ranking score is underlined. From Table 1, it can be observed that DecoratingFusion achieves the highest mAP across all three categories, outperforming all nine representative methods. Although DecoratingFusion does not achieve the best performance in the easy and moderate difficulty levels for car detection, it ranks first in the difficult difficulty level. For small objects (pedestrians and bicycles) detection, DecoratingFusion achieves the top rank across various difficulty levels.

\begin{table}
\caption{Performance comparison on the Waymo \emph{val} set for 3D vehicle (IoU = 0.7) and pedestrian (IoU = 0.5) detection.}
\centering
\begin{tabular}{c|c|c|c|c|c|c}
\hline
\multirow{2}*{Method} & \multirow{2}*{Modality} & mAPH & \multicolumn{2}{c|}{Vehicle(AP/APH)} & \multicolumn{2}{c}{Pedestrian(AP/APH)}\\
\cline{3-7}
& & L2 & L1 & L2 & L1 & L2\\
\hline %[2pt]
SECOND\cite{yan2018second} & L & 57.32 & 72.27/71.69 & 63.85/63.33 & 68.70/58.18 & 60.72/51.31\\
3D-MAN\cite{yang20213d} & L & 63.05 & 74.50/74.00 & 67.60/67.10 & 71.70/67.70 & 62.60/59.00\\
Part-A$^2$\cite{shi2020points} & L & 63.30 & 77.10/76.50 & 68.50/68.00 & 75.20/66.90 & 66.20/58.60\\
PDV\cite{hu2022point} & L & 63.55 & 76.85/76.33 & 69.30/68.81 & 74.19/65.96 & 65.85/58.28\\
CenterPoint\cite{yin2021center} & L & 64.40 & - & -/66.20 & - & -/62.60\\
Centerformer\cite{zhou2022centerformer} & L & \underline{74.40} & 78.80/78.30 & \textbf{74.30/73.80} & 82.10/79.30 & 77.80/75.00\\
\hline
PointAugmenting\cite{wang2021pointaugmenting} & L+C & 63.40 & 67.40/- & 62.70/62.20 & 75.04/- & 70.60/64.60\\
DeepFusion\cite{li2022deepfusion} & L+C & 74.20 & \underline{80.60/80.10} & 72.90/72.40 & \underline{85.80/83.00} & \underline{78.70/76.00}\\
\hline
DecoratingFusion & L+C & \textbf{74.80} & \textbf{81.52/81.08} & \underline{73.74/73.11} & \textbf{86.21}/\textbf{83.65} & \textbf{79.15}/\textbf{76.49}\\
\hline
\end{tabular}
\end{table}

\subsubsection{Waymo.}
On the larger and more diverse Waymo dataset, we also compare DecoratingFusion with several state-of-the-art methods, and the experimental results are presented in Table 2. It can be observed that on Waymo's official primary difficulty metric, L2, DecoratingFusion achieves first place in pedestrian detection and is only 0.69\% behind the top-ranking method, CenterFormer, in vehicle detection. Although CenterFormer, a lidar-only method, achieves the top rank in vehicle detection at L2 difficulty, its performance in pedestrian detection is not remarkable. This is due to the inherent limitation of lidar-only methods, as point cloud data itself is sparse in 3D space and has limited coverage of small object instances. In terms of L1 difficulty, DecoratingFusion not only achieves first place in pedestrian detection but also attains the best performance in vehicle detection, surpassing the second-ranking method, DeepFusion, by 0.98\%. Overall, considering all categories, DecoratingFusion secures the first rank in L2 difficulty mAPH. Similar to its performance on the KITTI dataset, DecoratingFusion demonstrates advantages in small object recognition on the Waymo dataset as well.

\subsection{Ablation study}
To demonstrate the effectiveness of each component in DecoratingFusion, we conduct two sets of experiments on the Waymo dataset, specifically on the L2 difficulty level. These experiments focused on evaluating the point-level fusion module and the feature fusion module separately.

\begin{table}
\caption{Effect of each component in the point-level fusion module.}
\centering
\begin{tabular}{ccc|cc}
\hline
Decoration & E2E & 2SparseConv & Vehicle & Pedestrian\\
\hline %[2pt]
& & & 72.40 & 76.00\\
\ding{51} & & & 72.84(+0.44) & 76.35(+0.35)\\
\ding{51} & \ding{51} & & 72.93(+0.53) & 76.4(+0.40)\\
\ding{51} & & \ding{51} & 72.87(+0.47) & 76.38(+0.38)\\
\ding{51} & \ding{51} & \ding{51} & 72.97(+0.57) & 76.42(+0.42)\\
\hline
\end{tabular}
\end{table}

The point-level fusion module consists of three components: Decoration, E2E, and 2SparseConv. Decoration indicates whether to use image features to decorate the original point cloud data, E2E represents whether to train the 2D network independently or in an end-to-end manner with other network components, and 2SparseConv indicates whether to use two separate sparse convolutions to extract the decorated point cloud data. In this set of experiments, the baseline method chosen is DeepFusion, and the results are shown in Table 3. From the results, it can be observed that Decoration brings the largest improvement, indicating that the hard correlation between point cloud and image can significantly enhance the feature fusion. Additionally, both the E2E and 2SparseConv modules contribute positively to the model's performance, with E2E providing a relatively larger improvement.

\begin{table}
\caption{Effect of each component in the feature fusion module.}
\centering
\begin{tabular}{cc|cc}
\hline
Heatmap Init. & Category Embedding & Vehicle & Pedestrian\\
\hline %[2pt]
& & 72.84 & 76.35\\
\ding{51} & & 72.97(+0.13) & 76.41(+0.06)\\
\ding{51} & \ding{51} & 73.02(+0.18) & 76.43(+0.08)\\
\hline
\end{tabular}
\end{table}

The feature fusion module consists of two components: HeatmapInit. (which initializes the query with a center point heatmap) and Category Embedding (which embeds category information into the query). It is important to note that the Category Embedding component relies on the HeatmapInit. component, as the category information is derived from the predictions of the center heatmap. The baseline method for this set of experiments is DeepFusion + Decoration. From Table 4, it can be seen that both modules contribute positively to the model's performance, with the main improvement coming from the HeatmapInit. component.

\section{Conclusion}
We combine the early fusion and mid-level fusion in multi-modal fusion methods and propose a new 3D object detection network called DecoratingFusion. The core idea of DecoratingFusion is to establish a hard correlation between point cloud and image using calibration matrices. It utilizes the decorated point cloud features to guide the generation of object queries and finally fuses the point cloud and image features through cross-attention mechanisms. DecoratingFusion consists of two stages: point-level fusion and feature fusion. In the point-level fusion stage, instead of using independently pre-trained networks, image features are learned in an end-to-end manner. Additionally, two separate sparse convolutions are used to extract the decorated point cloud features. In the feature fusion stage, the object query is initialized with a center point heatmap, which brings the initial position of the query closer to the actual center of the object. Moreover, the predicted category from the center point heatmap is embedded as supplementary information into the object query. Experiments conducted on the KITTI and Waymo datasets demonstrate the superiority of DecoratingFusion in 3D object detection.


\bibliographystyle{splncs04}
\bibliography{mybibliography}

\begin{thebibliography}{10}
\providecommand{\url}[1]{\texttt{#1}}
\providecommand{\urlprefix}{URL }
\providecommand{\doi}[1]{https://doi.org/#1}

\bibitem{bai2022transfusion}
Bai, X., Hu, Z., Zhu, X., Huang, Q., Chen, Y.: Transfusion: Robust lidar-camera
  fusion for 3d object detection with transformers. In: Proceedings of the
  IEEE/CVF Conference on Computer Vision and Pattern Recognition. pp.
  1090--1099 (2022)

\bibitem{chen2022focal}
Chen, Y., Li, Y., Zhang, X., Sun, J., Jia, J.: Focal sparse convolutional
  networks for 3d object detection. In: Proceedings of the IEEE/CVF Conference
  on Computer Vision and Pattern Recognition. pp. 5428--5437 (2022)

\bibitem{hu2022point}
Hu, J.S., Kuai, T., Waslander, S.L.: Point density-aware voxels for lidar 3d
  object detection. In: Proceedings of the IEEE/CVF Conference on Computer
  Vision and Pattern Recognition. pp. 8469--8478 (2022)

\bibitem{huang2020epnet}
Huang, T., Liu, Z., Chen, X., Bai, X.: Epnet: Enhancing point features with
  image semantics for 3d object detection. In: Proceedings of the European
  conference on computer vision (ECCV). pp. 35--52. Springer (2020)

\bibitem{li2022deepfusion}
Li, Y., Yu, A.W., Meng, T., Caine, B., Ngiam, J.: Deepfusion: Lidar-camera deep
  fusion for multi-modal 3d object detection. In: Proceedings of the IEEE/CVF
  Conference on Computer Vision and Pattern Recognition. pp. 17182--17191
  (2022)

\bibitem{pang2020clocs}
Pang, S., Morris, D., Radha, H.: Clocs: Camera-lidar object candidates fusion
  for 3d object detection. In: 2020 IEEE/RSJ International Conference on
  Intelligent Robots and Systems (IROS). pp. 10386--10393. IEEE (2020)

\bibitem{qi2018frustum}
Qi, C.R., Liu, W., Wu, C., Su, H., Guibas, L.J.: Frustum pointnets for 3d
  object detection from rgb-d data. In: Proceedings of the IEEE conference on
  computer vision and pattern recognition. pp. 918--927 (2018)

\bibitem{shi2020pv}
Shi, S., Guo, C., Jiang, L., Wang, Z., Shi, J., Wang, X., Li, H.: Pv-rcnn:
  Point-voxel feature set abstraction for 3d object detection. In: Proceedings
  of the IEEE/CVF conference on computer vision and pattern recognition. pp.
  10529--10538 (2020)

\bibitem{shi2019pointrcnn}
Shi, S., Wang, X., Li, H.: Pointrcnn: 3d object proposal generation and
  detection from point cloud. In: Proceedings of the IEEE/CVF conference on
  computer vision and pattern recognition. pp. 770--779 (2019)

\bibitem{shi2020points}
Shi, S., Wang, Z., Shi, J., Wang, X., Li, H.: From points to parts: 3d object
  detection from point cloud with part-aware and part-aggregation network. IEEE
  transactions on pattern analysis and machine intelligence  \textbf{43}(8),
  2647--2664 (2020)

\bibitem{vora2020pointpainting}
Vora, S., Lang, A.H., Helou, B., Beijbom, O.: Pointpainting: Sequential fusion
  for 3d object detection. In: Proceedings of the IEEE/CVF conference on
  computer vision and pattern recognition. pp. 4604--4612 (2020)

\bibitem{wang2021pointaugmenting}
Wang, C., Ma, C., Zhu, M., Yang, X.: Pointaugmenting: Cross-modal augmentation
  for 3d object detection. In: Proceedings of the IEEE/CVF Conference on
  Computer Vision and Pattern Recognition. pp. 11794--11803 (2021)

\bibitem{yan2018second}
Yan, Y., Mao, Y., Li, B.: Second: Sparsely embedded convolutional detection.
  Sensors  \textbf{18}(10), ~3337 (2018)

\bibitem{yang20213d}
Yang, Z., Zhou, Y., Chen, Z., Ngiam, J.: 3d-man: 3d multi-frame attention
  network for object detection. In: Proceedings of the IEEE/CVF conference on
  computer vision and pattern recognition. pp. 1863--1872 (2021)

\bibitem{yin2021center}
Yin, T., Zhou, X., Krahenbuhl, P.: Center-based 3d object detection and
  tracking. In: Proceedings of the IEEE/CVF conference on computer vision and
  pattern recognition. pp. 11784--11793 (2021)

\bibitem{zhang2022cat}
Zhang, Y., Chen, J., Huang, D.: Cat-det: Contrastively augmented transformer
  for multi-modal 3d object detection. In: Proceedings of the IEEE/CVF
  Conference on Computer Vision and Pattern Recognition. pp. 908--917 (2022)

\bibitem{zheng2021se}
Zheng, W., Tang, W., Jiang, L., Fu, C.W.: Se-ssd: Self-ensembling single-stage
  object detector from point cloud. In: Proceedings of the IEEE/CVF Conference
  on Computer Vision and Pattern Recognition. pp. 14494--14503 (2021)

\bibitem{zhou2019objects}
Zhou, X., Wang, D., Kr{\"a}henb{\"u}hl, P.: Objects as points. arXiv preprint
  arXiv:1904.07850  (2019)

\bibitem{zhou2022centerformer}
Zhou, Z., Zhao, X., Wang, Y., Wang, P., Foroosh, H.: Centerformer: Center-based
  transformer for 3d object detection. In: European Conference on Computer
  Vision. pp. 496--513. Springer (2022)

\end{thebibliography}





% LNCS Homepage, \url{http://www.springer.com/lncs}, last accessed 2023/10/25
% \end{thebibliography}
\end{document}